\documentclass{article}
\usepackage[preprint, nonatbib]{nips_2018}
\usepackage[utf8]{inputenc} 
\usepackage[T1]{fontenc}    
\usepackage{hyperref}
\usepackage{url}
\usepackage{booktabs}
\usepackage{amsfonts}
\usepackage{nicefrac}
\usepackage{microtype}
\usepackage{graphicx}
\usepackage{multirow}

\title{Automated Image Data Preprocessing with Deep Reinforcement Learning}

\author{
 Tran Ngoc Minh\textsuperscript{1},  Mathieu Sinn\textsuperscript{2},  Hoang Thanh Lam\textsuperscript{3},  Martin Wistuba\textsuperscript{4} \\
 IBM Research Dublin, Ireland \\
 \textsuperscript{1,4}\texttt{\{m.n.tran, martin.wistuba\}@ibm.com},  \textsuperscript{2,3}\texttt{\{mathsinn, t.l.hoang\}@ie.ibm.com} \\
}

\begin{document}
\maketitle

\begin{abstract}
Data preparation, i.e.~the process of transforming raw data into a format that can be used for training effective machine learning models, is a tedious and time-consuming task. For image data, preprocessing typically involves a sequence of basic transformations such as cropping, filtering, rotating or flipping images. Currently, data scientists decide manually based on their experience which transformations to apply in which particular order to a given image data set. Besides constituting a bottleneck in real-world data science projects, manual image data preprocessing may yield suboptimal results as data scientists need to rely on intuition or trial-and-error approaches when exploring the space of possible image transformations and thus might not be able to discover the most effective ones. To mitigate the inefficiency and potential ineffectiveness of manual data preprocessing, this paper proposes a deep reinforcement learning framework to automatically discover the optimal data preprocessing steps for training an image classifier. The framework takes as input sets of labeled images and predefined preprocessing transformations. It jointly learns the classifier and the optimal preprocessing transformations for individual images. Experimental results show that the proposed approach not only improves the accuracy of image classifiers, but also makes them substantially more robust to noisy inputs at test time.
\end{abstract}

\section{Introduction}
\label{sec1}
Data preprocessing, i.e.~the process of transforming raw data into a format that can be used for training effective machine learning models, accounts for 50-80\% of the time spent on typical data science projects \cite{bib1, bib2}. Besides constituting a bottleneck, manual data preprocessing is also ineffective as it only explores a small part of the space of possible transformations and thus might not discover the most effective ones for removing noise and/or extracting meaningful features from a given set of raw data. Unstructured data\footnote{By \textit{unstructured data} we mean images, text and time series, while we use \textit{structured data} to refer to data in tabular format, e.g.~as in relational databases.} are particularly challenging in this regard as their preparation requires deep expertise in fields such as Computer Vision or Natural Language Preprocessing; moreover, because of the high complexity of machine learning models dealing with such data, the effect of data preprocessing is particularly difficult to understand. Hence, automating data preprocessing is highly desirable as it increases the productivity of data scientists and may lead to better performance of the resulting machine learning models.

Despite of its high potential value, the automation of data preprocessing has been mostly overlooked by the machine learning community, with only few prior works on this subject \cite{bib1, bib3}. Recently, Bilalli et al. \cite{bib1} suggested a method for automating data preprocessing via meta-learning. However, their approach only focuses on structured data with a limited number of relatively simple preprocessing techniques such as normalization, standardization and discretization. Furthermore, preprocessing in their case does not address individual data instances, but instead applies to the whole data set. The study in \cite{bib3} develops a transformation pursuit for image classification, which is also applied to the whole data set. While these studies provide principled methods for augmenting training data, they still suffer from a number of intrinsic disadvantages. Firstly, given a large set of transformations\footnote{Throughout this paper, we use the words \textit{preprocessing} and \textit{transformation} interchangeably to indicate an operation applied to data instances such as flipping an image.}, applying them to all data instances results in an excessively large training data set, increasing prohibitively the amount of training time. Secondly, note that the order of applying transformations is important; for example, rotating and then flipping an image creates a different result than first flipping and then rotating the image. With a large transformation set, finding effective chains of transformations\footnote{The term \textit{chain of transformations} is used to indicate an ordered set of transformations.} requires an exhaustive search over an exponential search space. Thirdly, applying the same transformation chain to all data instances is often ineffective as each data instance has its own feature and hence should be preprocessed in a different way. For example, different images might require rotations with different angles because they were taken from different views; clearly, applying the same rotation to them in the augmented training set is inefficient. Lastly, an augmentation with irrelevant transformations may even produce wrongly labeled data instances, e.g. a 180-degree rotation of an image of the digit ``6'' produces an image of the digit ``9''. 

In this work, we propose an automated method for data preprocessing that transforms each data instance individually\footnote{An implementation of the method can be found at \href{https://github.com/IBM/automation-of-image-data-preprocessing}{\it{https://github.com/IBM/automation-of-image-data-preprocessing}}.}. Although we present our approach for image inputs, it easily generalizes to other types of data, such as time series or text. The proposed approach is based on a deep reinforcement learning framework to address the limitations of state-of-the-art approaches discussed above: Firstly, our preprocessing approach results in one variant per image, so the augmented training data set has the same size as the original one. Secondly and thirdly,  we preprocess each image individually with its own transformation chain which is discovered on-the-fly, using reinforcement learning to explore the space of potential transformation chains. Lastly, our approach discovers optimal transformation chains, again by using reinforcement learning to exploit the most effective ones. Experimental results on real-world data sets show that the more noise a data set contains, the more effective the framework is to improve the accuracy of a classifier trained on that data set. Furthermore, the classifier also becomes substantially more robust against noisy inputs at test time when being trained using our proposed framework.

The remainder of this paper is organized as follows. We discuss related work in Section \ref{sec2} and then present the addressed challenge as well as our proposed solution in Section \ref{sec3}. In order to evaluate the solution, we describe an evaluation methodology in Section \ref{sec4} and experiments in Section \ref{sec5}. Finally, we discuss and conclude our study in Sections \ref{sec6} and \ref{sec7}.

\section{Related Work}
\label{sec2}
Generalization is the main challenge of image classifiers, particularly when trained on small and/or noisy training data sets. Therefore, numerous approaches have been proposed to improve the generalization, such as adding a regularization term on the norm of weights \cite{bib4}, using dropout \cite{bib5, bib6} or batch normalization \cite{bib7}. Data augmentation is another effective approach that helps increase the generalization of image classifiers through applying simple transformations such as rotating and flipping input images, and adding the transformed images to the training set \cite{bib8}. The full set of transformations used in \cite{bib8} includes shifting, zooming in/out, rotating, flipping, distorting, shading and styling. Data augmentation with more complicated transformations is investigated in \cite{bib9}, which evaluates three concrete preprocessing techniques, namely Zero Component Analysis, Mean Normalization and Standardization, on the performance of different convolutional neural networks. While the approaches in \cite{bib9, bib8} preprocess images according to a preselected chain of transformations, Paulin et al. \cite{bib3} suggest that the transformation set should be chosen in principled way instead of resorting to (manual) trial-and-error, which is feasible only when the number of possible transformations is small. Their proposed approach selects a set of transformations, possibly ordered, through a greedy search strategy. Although this approach offers a more competitive set of transformations, it still has several limitations: Firstly, the search process is inefficient because it involves retraining the classifier on the whole augmented data set every time a candidate transformation in the search space is evaluated. Secondly, the same preprocessing transformations are applied to all images, which has a number of disadvantages as discussed in Section \ref{sec1}. Our approach uses a reinforcement learning framework to address precisely those shortcomings. 

Reinforcement learning \cite{bib10} and specifically deep reinforcement learning \cite{bib11} have recently drawn substantial attention from the machine learning research community. However, there are only few studies \cite{bib13, bib14, bib12} applying deep reinforcement learning to visual recognition tasks, such as edge detection, segmentation and object detection, or active object localization. However, none of these works considers automating the preprocessing of images or learning transformation sets. To the best of our knowledge, our work is the first study utilizing deep reinforcement learning to search for effective chains of preprocessing transformations for individual images.

\section{Reinforcement Learning Framework}
\label{sec3}
\begin{figure}[ht]
\centering
\includegraphics[width=13cm,height=4.5cm]{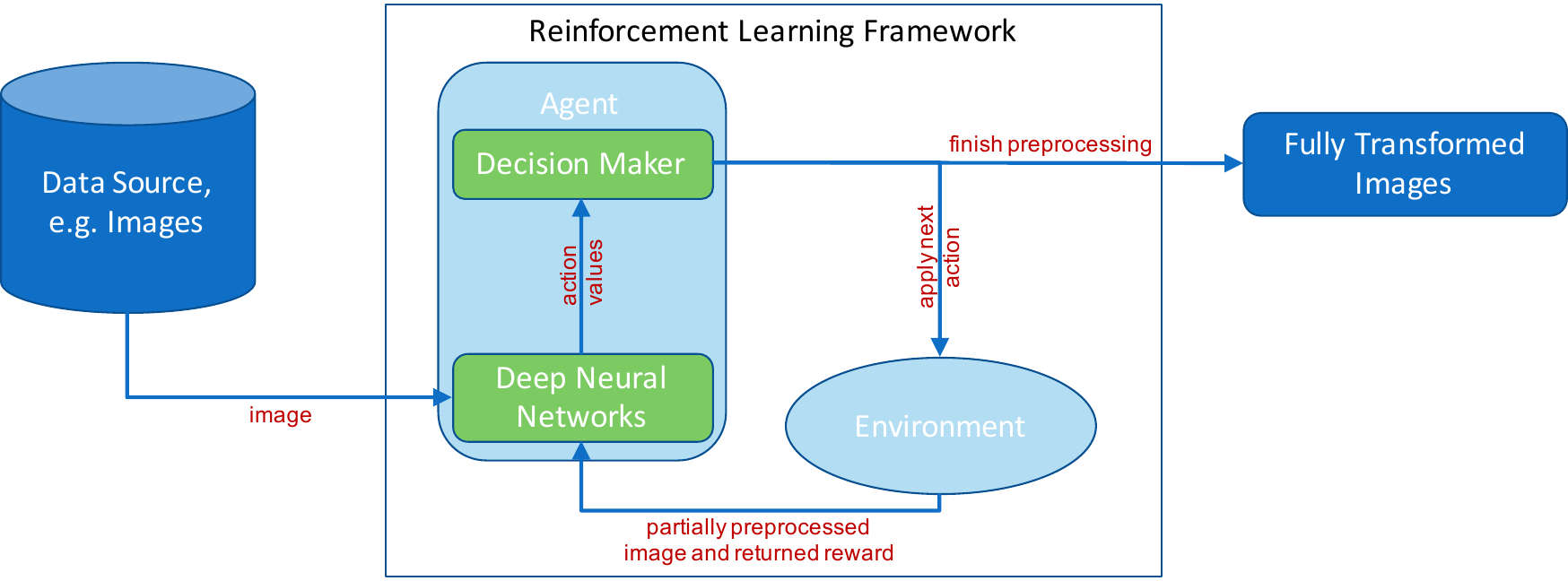}
\caption{The workflow of preprocessing images.}
\label{fig1}
\end{figure}

We present in this section our detailed solution for image preprocessing using a deep reinforcement learning framework. The framework includes two basic components, namely, an agent and an environment. Given an input image, the general workflow of the framework is shown in Figure \ref{fig1}. The agent has its policy represented by a deep neural network whose outputs are action values used by a decision maker to decide whether an image is sufficiently preprocessed or  whether additional transformations need to be applied. The environment is responsible for transforming the image upon request by the agent, where a concrete transformation operation is given together with the request. After doing so, the environment continues to feed the transformed image into the deep neural network for a new evaluation. This process is iterated until the agent makes a final decision to stop preprocessing the image. In addition to performing the actual image transformations, the environment also evaluates the impact of each transformation and produces rewards based on that. Technical details of each component of the framework are given in the following subsections.

\subsection{State and Action Spaces}
In this section we define states and actions used in our reinforcement learning framework and introduce an important property of preprocessing techniques that can be used.

\subsubsection{State Space}
We define a state as an image, thus an original state corresponds to an original image, and a transformed state corresponds to a preprocessed image. As such, the state space consists of all original images as well as all images transformed by an arbitrary chain of transformations. It is easy to see that the size of the state space grows exponentially in the length of the transformation chains.

\subsubsection{Action Space}
\label{actspace}
\begin{figure}[ht]
\centering
\includegraphics[width=13cm,height=4.5cm]{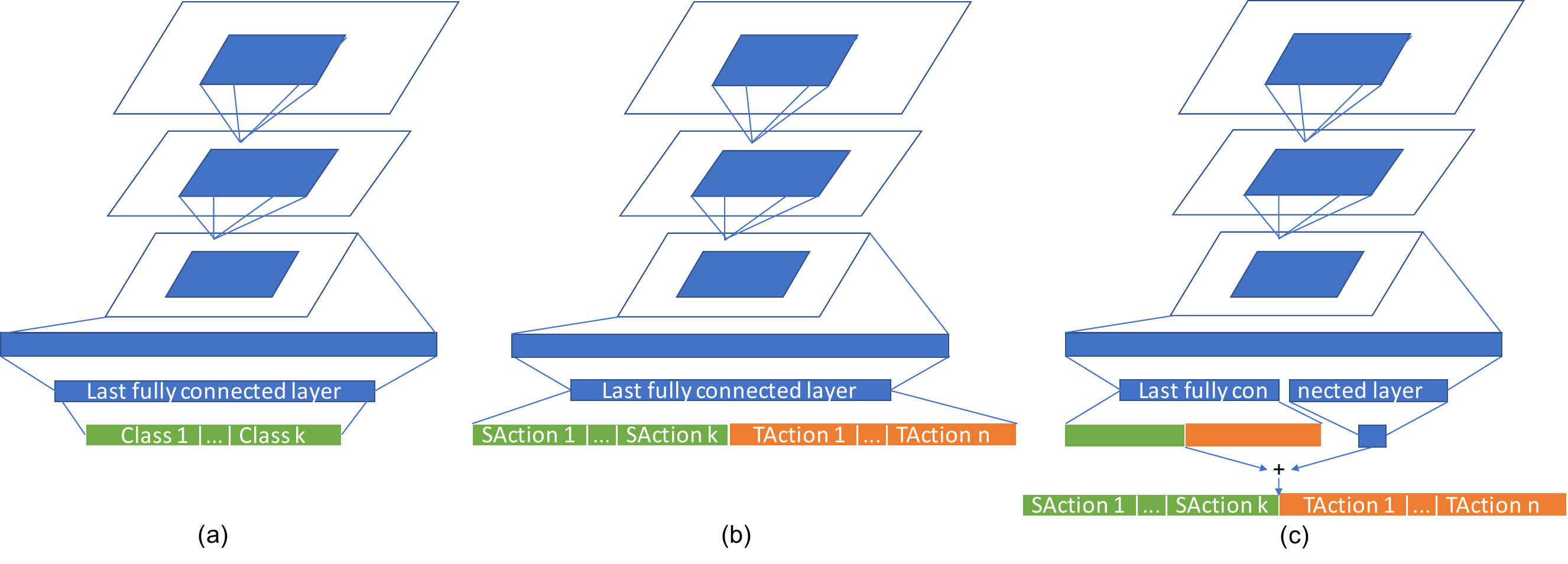}
\caption{Illustration of deep neural network policies}.
\label{fig2}
\end{figure}

In Figure \ref{fig2}(a), we show the typical architecture of a convolutional neural network (CNN) whose output is a logit vector with $k$ values representing the corresponding unnormalized likelihood of the $k$ output classes. In our study, we use a variant of Deep Q-Network (DQN) \cite{bib15, bib16} to model a network policy. The network policy implemented in a DQN as shown in Figure \ref{fig2}(b) resembles the CNN in Figure \ref{fig2}(a), except that the output layer is extended to form an action space. The DQN output layer containing Q-values consists of two parts that are corresponding to two groups of actions. The first part is a vector playing the same role as the logit vector in the CNN, i.e. it represents the unnormalized likelihood of the $k$ classes. We denote each slot in this part as a stop action $SAction_i$. If the decision maker decides on the next action as one of the stop actions $SAction_i$, the preprocessing of an input image will stop with a prediction of class $i$ for the input image. The second part of the DQN output layer is a vector representing a set of $n$ transformation actions. If a decision is made for a next action with one of the transformation actions $TAction_j$, the current image will be continued to be preprocessed with the transformation $j$. The two sets of stop and transformation actions form an action space which has totally $k+n$ actions in case of discrete transformation. Note that it is straightforward to also support continuous actions. For example, we can model a continuous rotation by defining two slots in the second part of the DQN output: one for the Q-value of the rotation action and one for the value of the rotation angle. Likewise, we can also adapt the first part of the DQN output in order to apply the framework to a regression problem, e.g.~when the inputs are time series and the task is to forecast future values.

\subsubsection{Symmetry of a Transformation Action}
\label{sym}
There is an important property that any considered transformation operation should possess, the so-called symmetry property: Due to the exploratory nature of the reinforcement learning approach, it is essential to allow an image to be recovered after a trial transformation that resulted in a poor performance. For example, if there is a rotation transformation with an angle $\alpha$, there should be another rotation transformation with the angle $-\alpha$ in the set of possible transformations. For transformations such as image cropping, inverting transformations is not as straight-forward; it requires implementing a memory mechanism to remember states before any transformation. As this may lead to large memory usage for long transformations, a mechanism to control the maximum length of a transformation chain should be used (see Section \ref{env}).

\subsection{Decision Maker}
The decision maker is where a reinforcement learning policy is deployed. It is responsible for selecting the next action to be applied to the current state. The action and the state are then passed to the environment component for further processing. In our study, we use the max policy to select an appropriate action, given the DQN output layer. Furthermore, in order to enable the exploration of reinforcement learning, we allow the decision maker to select alternative next actions randomly with some probability $\epsilon$, which is known as $\epsilon$-greedy exploration strategy \cite{bib10}. The probability $\epsilon$ starts at a maximum of $1.0$ and is annealed down to a minimum of $0.1$ during training. 

\subsection{Deep Neural Network}
Using a DQN as in Figure \ref{fig2}(b) is a simple starting point; performance gains can be achieved using other variants of DQN. In our work, we implemented a variant of DQN, namely Dueling DQN (DDQN) \cite{bib17} as shown in Figure \ref{fig2}(c). The idea behind DDQNs is that the Q-values will be a combination of a value function and an advantage function. The value function specifies how good it is to be in a given state while the advantage function indicates how much better selecting an action is compared to the others. The benefit of separating the two functions is that the reinforcement learning framework does not need to learn both value and advantage at the same time, and therefore a DDQN is able to learn the state-value function efficiently. In order to update the deep neural network, we use the Bellman equation $Q(s,a) = r + \gamma \times max_{a'}(Q(s',a'))$, where $Q(s,a)$ is the DQN output value of action $a$ given input state $s$, $r$ and $s'$ are the reward and the next state returned by the environment when the action $a$ is applied to the state $s$, and $\gamma$ is the discounted parameter. We refer the reader to \cite{bib17} for more details on DDQNs.
 
\subsection{Environment}
\label{env}
The environment is where actual transformations on images are performed. In addition, it is also responsible for calculating return rewards during training. Upon receiving an image and an action from the reinforcement learning agent, the environment behaves differently depending on the type of the action. If it is a transformation action, the environment will apply that transformation to the image only if the length of the chain of transformations applied particularly to that image is smaller than a configurable parameter $max\_len$. If the chain is longer than $max\_len$, the image is recovered to its original state and the reinforcement learning framework must seek another transformation chain for the image. Note, this recovery mechanism is only used for training. At test time, we simply pick the stop action with the largest Q-value for the prediction of the image. The recovery mechanism also solves the memory problem described in Section \ref{sym}. Regardless of the length of the current transformation chain, the environment will return a zero reward to the reinforcement learning agent in this case. 

If the environment receives a stop action $SAction_i$, it does not return a new image but a reward and classifies the original image as class $i$. The strategy to compute return rewards during training plays an important role for the convergence of the training. The environment uses the ground true label of the original image to determine the reward. A simple strategy is to assign a reward of $+1$ if the label is equal to $i$ and $-1$ if otherwise. However, this simple strategy does not work well when the number of classes $k$ is larger than 2 since it causes unbalancing rewards. Hence, we suggest a more robust scheme to compute return rewards, which is to assign a reward of $k-1$ if the label is equal to $i$ and $-1$ if otherwise.

\begin{figure}[ht]
\centering
\includegraphics[width=12cm,height=5cm]{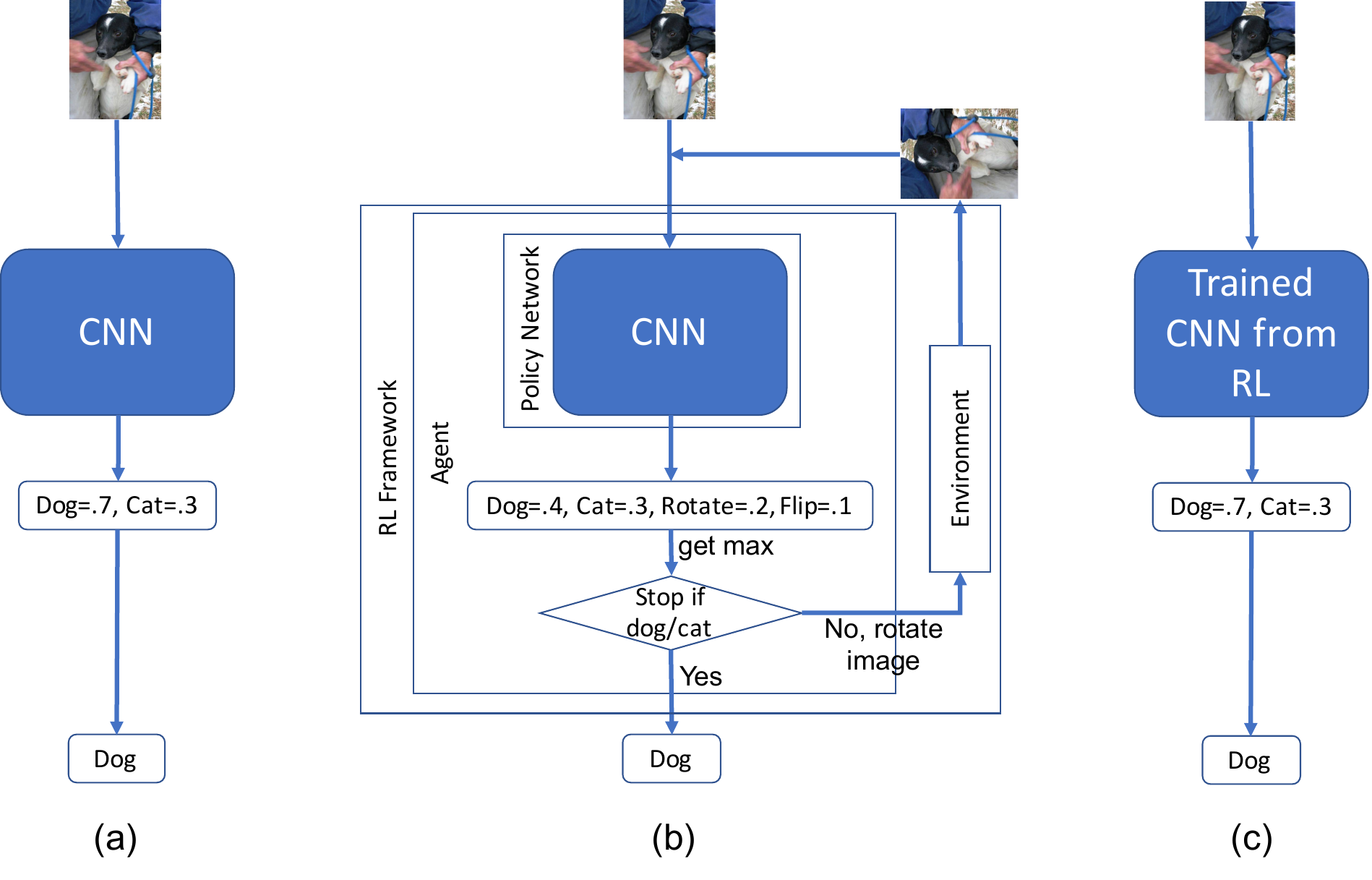}
\caption{The methodology of our experiments.}
\label{fig3}
\end{figure}

\section{Methodology}
\label{sec4}
Our methodology in setting up experiments is illustrated in Figure \ref{fig3}. In order to evaluate our auto-preprocessing framework, we train three different models, namely NN, RL and CL, shown in Figures \ref{fig3}(a), \ref{fig3}(b) and \ref{fig3}(c), respectively. For all of them, we use the same neural network architecture. Figure \ref{fig3}(a) represents a CNN model with an arbitrary architecture. This same architecture will also be used as the policy network in the reinforcement learning framework as shown in Figure \ref{fig3}(b). Since both models use the same network architecture, any performance difference between the two models in our experiments is caused by the reinforcement learning solution. In Figure \ref{fig3}(c), we also have a CNN model with the same network architecture, but we do not train the network from scratch. Rather, we continue to fine-tune the network obtained from the reinforcement learning framework. The $N$ original training images are preprocessed by the framework to produce $N$ new training images which are used as inputs of the fine-tuning process.

\begin{figure}[ht]
\centering
\includegraphics[width=13cm,height=4.5cm]{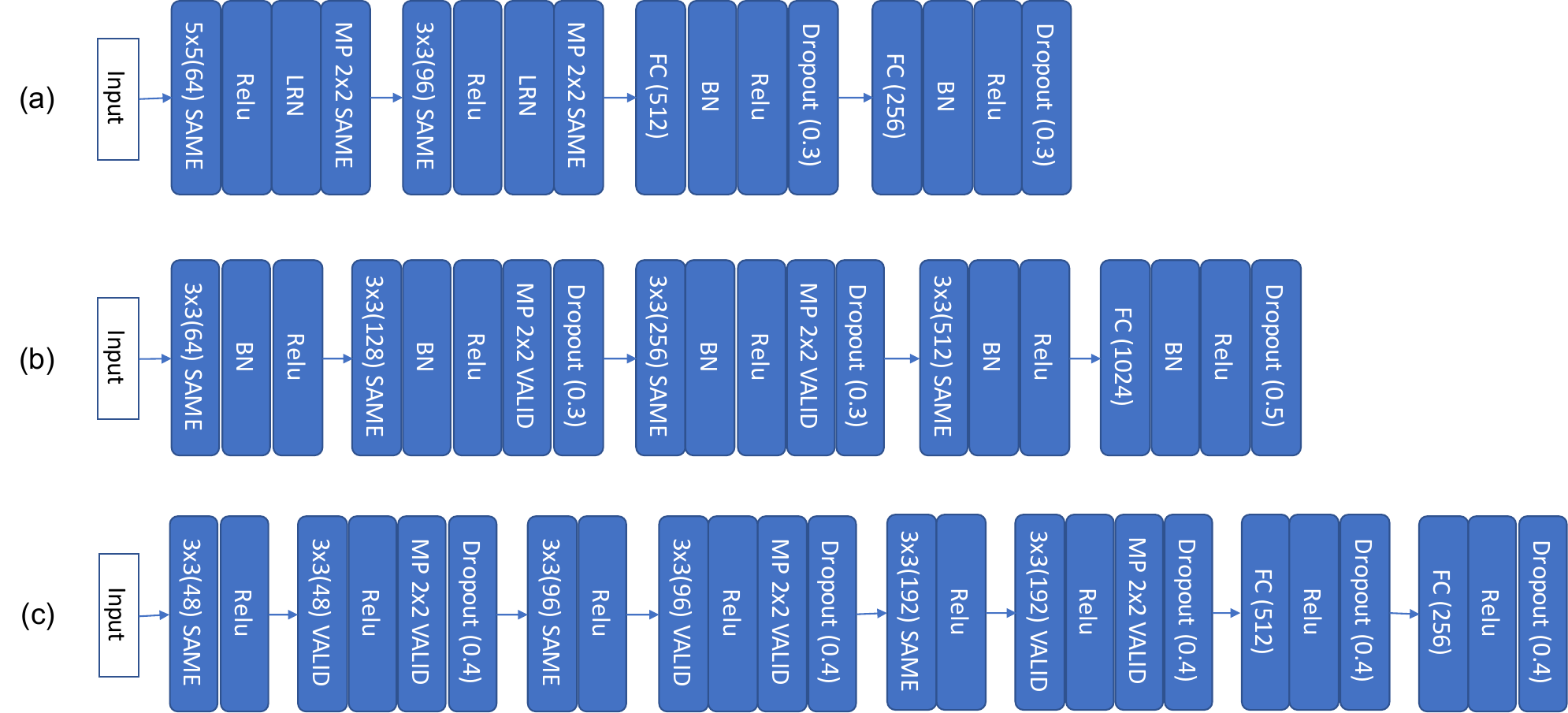}
\caption{CNN architectures used in our experiments. LRN, BN, MP and FC stand for local response normalization, batch normalization, max pooling and fully connected, respectively. All convolutional layers use a stride of 1x1 and all max pooling layers use a stride of 2x2.}
\label{fig4}
\end{figure}

In our experiments, we implement three different CNN architectures, namely Arch1, Arch2 and Arch3 as shown in Figures \ref{fig4}(a), \ref{fig4}(b) and \ref{fig4}(c), respectively. Hence, we have totally nine models for comparison. The architectures are selected according to their complexity ranging from simple in Figure \ref{fig4}(a) to complex in Figure \ref{fig4}(c). Note that the hyperparameters and architectures of the models in Figure \ref{fig4} are not designed ``optimally'' (e.g.~using (hyper-)parameter tuning or auto-architecture search), but chosen such that there is some level of complexity difference between them, the effect of which is discussed in our evaluation below.

\section{Experimental Results}
\label{sec5}
In this section we will present our experiments to validate our solution to the problem of image preprocessing automation. We start our experiments by comparing the accuracy performance of image classifiers with and without preprocessing. Then, we evaluate the robustness of the classifiers with respect to distorted images at test time. In addition, we also provide some insights on the behaviour of the reinforcement learning framework. 

\subsection{Experimental Setups}
We select for our study four data sets with different levels of complexity and noise. MNIST \cite{bib18} is a very clean 10-class data set with 70K 28x28x1 images divided into 55K/5K/10K for training, validation and testing, respectively. SVHN \cite{bib19} is a 10-class data set that is noisier than MNIST with $\sim$864K 32x32x3 images divided into $\sim$598K/6K/26K. CIFAR \cite{bib20} is a 10-class data set that is yet noisier than SVHN with 60K 32x32x3 images divided into 45K/5K/10K. Finally, DOGCAT \cite{bib21} is the noisiest of all four data sets; it has 2 classes with 25K 100x100x3 images divided into 20K/1K/4K. 

With respect to the transformation set, we implemented two operations, namely image rotation and flipping, for simplicity because they trivially satisfy the symmetry property requirement without the necessity to implement a memory mechanism. Concretely, there are 11 transformations consisting of 3 flippings (horizontally, vertically, and both) and 8 rotations (with angles $-1, -2, -4, -8, +8, +4, +2, +1$ degrees). The parameter $max\_len$ specifying the maximum length of a transformation chain is set to $10$ in our experiments. Other general parameters include $optimizer=Adam$, $learning\_rate=0.0001$ and $regularization\_coefficient=0.001$. For each experiment, we perform 5 runs with 5 different initializations and report results as $mean \pm std$.

\subsection{Performance of Image Classifiers}
Performance results in terms of accuracy are shown in Table \ref{tableperf}. It can be seen that in most cases, the bare convolutional neural network classifier (NN) produces the worst performance while the reinforcement learning classifier (RL) yields higher accuracy. The accuracy performance is improved further by the CNN classifier that is continued to learn (CL) from the trained RL classifier. We note that the accuracy reported in Table \ref{tableperf} does not achieve state-of-the-art performance as the networks that we used in our experiments were relatively simple and not adapted for the data sets; nevertheless we believe it is worth mentioning that the RL framework results in improving the accuracy of the baseline methods, nb: without increasing the size of the training set. Moreover, it is interesting to observe that the accuracy difference between the NN classifier and the RL classifier increases for noisier and more complex data sets. On the one hand, while for MNIST simple preprocessing techniques such as rotation and flipping do not help improving the accuracy, they even decrease accuracy as some digits change their meaning when being rotated or flipped. On the other hand, on the much noisier DOGCAT data set, the RL classifier is much more successful in increasing the accuracy of the baseline CNN.

\begin{table}[ht]
\caption{Performance of image classifiers in term of accuracy.}
\centering
\begin{tabular}{| c  c | c | c | c | c |}
\hline 
&  & MNIST & SVHN & CIFAR & DOGCAT \\
\hline
\multirow{3}{*}{Arch1} & NN & $0.9926 \pm 0.0004$ & $0.9429 \pm 0.0015$ & $0.7018 \pm 0.0028$ & $0.7461 \pm 0.0137$ \\
& RL & $0.9915 \pm 0.0010$ & $0.9507 \pm 0.0031$ & $0.7442 \pm 0.0046$ & $0.8035 \pm 0.0062$ \\
& CL & $0.9935 \pm 0.0006$ & $0.9509 \pm 0.0026$ & $0.7410 \pm 0.0026$ & $0.8222 \pm 0.0102$   \\
\hline
\multirow{3}{*}{Arch2} & NN & $0.9941 \pm 0.0002$ & $0.9636 \pm 0.0008$ & $0.7910 \pm 0.0048$ & $0.8669 \pm 0.0261$ \\
& RL & $0.9916 \pm 0.0004$ & $0.9715 \pm 0.0016$ & $0.8193 \pm 0.0015$ & $0.9209 \pm 0.0124$ \\
& CL & $0.9936 \pm 0.0004$ & $0.9716 \pm 0.0011$ & $0.8175 \pm 0.0022$ & $0.9044 \pm 0.0090$ \\
\hline
\multirow{3}{*}{Arch3} & NN & $0.9954 \pm 0.0004$ & $0.9734 \pm 0.0016$ & $0.8391 \pm 0.0028$ & $0.8647 \pm 0.0065$ \\
& RL &  $0.9932 \pm 0.0007$ & $0.9726 \pm 0.0029$ & $0.8687 \pm 0.0080$ & $0.9231 \pm 0.0081$ \\
& CL & $0.9955 \pm 0.0005$ & $0.9760 \pm 0.0021$ & $0.8700 \pm 0.0057$ & $0.9243 \pm 0.0101$ \\
\hline
\end{tabular}
\label{tableperf}
\end{table}

\subsection{Robustness of Image Classifiers}
In order to evaluate the robustness of image classifiers, we distort each test image with $50\%$ probability by applying a random chain of transformations. Robustness results in term of accuracy are shown in Table \ref{tablerobust}. As we can see, the results are consistent in all cases in the sense that the NN classifier is less robust (its accuracy decreases significantly on the test set with distortions), compared to the performance on clean test data reported in Table \ref{tableperf}. On the other hand, the RL classifier is much more robust as its performance only slightly degrades on the distorted test data. Note that, only $50\%$ images of the test set were distorted, hence the robustness difference between the two classifiers would be even larger if all test images had been distorted. The robustness of the CL classifier is not as high as for the RL classifier, but still substantially higher than that of the NN classifier. This is a trade-off between the accuracy and the robustness when choosing between the RL and the CL classifiers. 

\begin{table}[ht]
\caption{Robustness of image classifiers in term of accuracy.}
\centering
\begin{tabular}{| c  c | c | c | c | c |}
\hline 
&  & MNIST & SVHN & CIFAR & DOGCAT \\
\hline
\multirow{3}{*}{Arch1} & NN & $0.7442 \pm 0.0037$ & $0.6991 \pm 0.0020$ & $0.5099 \pm 0.0078$ & $0.6687 \pm 0.0068$ \\
& RL & $0.9772 \pm 0.0024$ & $0.8699 \pm 0.0033$ & $0.7006 \pm 0.0085$ & $0.7574 \pm 0.0203$ \\
& CL & $0.8400 \pm 0.0048$ & $0.7945 \pm 0.0100$ & $0.6645 \pm 0.0032$ & $0.7723 \pm 0.0228$ \\
\hline
\multirow{3}{*}{Arch2} & NN & $0.7516 \pm 0.0021$ & $0.7242 \pm 0.0011$ & $0.5872 \pm 0.0057$ & $0.7936 \pm 0.0302$ \\
& RL & $0.9797 \pm 0.0016$ & $0.8876 \pm 0.0020$ & $0.7909 \pm 0.0027$ & $0.8987 \pm 0.0151$ \\
& CL & $0.8301 \pm 0.0114$ & $0.8207 \pm 0.0024$ & $0.7709 \pm 0.0035$ & $0.8524 \pm 0.0076$ \\
\hline
\multirow{3}{*}{Arch3} & NN & $0.7563 \pm 0.0024$ & $0.7400 \pm 0.0019$ & $0.6402 \pm 0.0093$ & $0.7791 \pm 0.0062$ \\
& RL & $0.9826 \pm 0.0014$ & $0.9062 \pm 0.0015$ & $0.8160 \pm 0.0046$ & $0.8872 \pm 0.0074$ \\
& CL & $0.9549 \pm 0.0046$ & $0.8462 \pm 0.0085$ & $0.7199 \pm 0.0037$ & $0.8845 \pm 0.0068$ \\
\hline
\end{tabular}
\label{tablerobust}
\end{table}

\subsection{A Deeper Look into the Operation of the Framework}
\begin{figure}[ht]
\centering
\includegraphics[width=12cm]{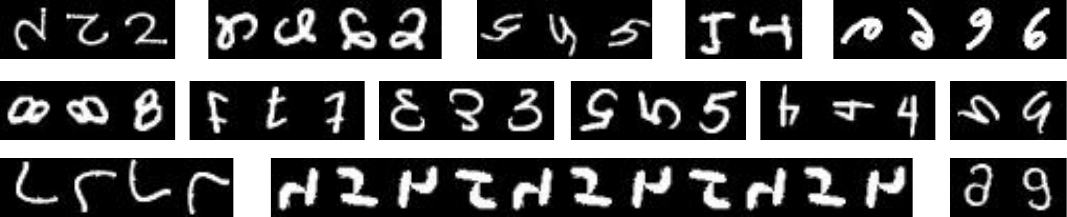}
\caption{Illustration of how the reinforcement learning framework preprocesses distorted images.}
\label{fig5}
\end{figure}

In order to visualize how the reinforcement learning framework preprocesses distorted images, we run another experiment on MNIST with coarser distortions, in particular rotations are performed with large angles $\pm90$ degrees and flipping operations as in the previous experiment. For each distorted image, we trace the operation of the framework and obtain the transformation chain that the framework automatically generates for the image. An illustration for a few images is shown in Figure \ref{fig5}. It is interesting that most images are either classified directly or transformed to their original version before being classified. The exact recovery is possible thanks to the symmetry property of transformation actions. Although the framework is able to recover distorted images, it is not guaranteed to find the optimal chain of transformations in term of the shortest recovery path. In addition, there is a small number of images which are confused by the framework as shown in the bottom row of Figure \ref{fig5}. These are the main source of misclassification errors of the reinforcement learning classifier. 

\section{Discussion}
\label{sec6}
The key contributions of this paper are three-fold. Firstly, we developed the idea of automated data preprocessing using a reinforcement learning framework. While we demonstrated and evaluated it for image data, it is applicable to other types of structured and unstructured data as well. Secondly, the proposed system is iterative and therefore it provides explainable data preprocessing, i.e. one can inspect which transformations were applied to each data instance during the preprocessing. Thirdly, compared with traditional data augmentation approaches, our system follows a more efficient approach to produce a clean training data set that can be used effectively for training highly accurate and robust machine learning models.

Despite being of high practical value, the automation of data preprocessing has only drawn little interest by the machine learning research community so far. Although we suggest in this paper a novel approach for this problem, there is still a lot of room to extend this work. Firstly, the set of transformations may contain more advanced preprocessing techniques such as rotations with learnable angles, cropping/scaling with learnable ratios, image segmentation, object detection, etc. While it is easy to integrate continuous actions with learnable parameters into the framework as described in Section \ref{actspace}, complicated actions like image segmentation and object detection may require more efforts. For example, one could select only a small number of segments or objects as the simplified representation of an image for the next iteration after applying those actions. Secondly, one could boost the performance of the reinforcement learning framework by replacing the current simple DQN policy network. In addition, CNNs derived from the policy network (as described in Figure \ref{fig3}(c)) may be a way to obtain better performance in terms of accuracy.

\section{Conclusions}
\label{sec7}
We have presented in this paper a novel approach to the problem of automating data preprocessing, which is of high potential value for real-world data science and machine learning projects. The approach is based on a reinforcement learning framework to find sequences of preprocessing transformations for  each data instance individually. We showed in our experiments that even with simple preprocessing actions such as rotation and flipping, image classifiers can benefit significantly with respect to their accuracy and particularly their robustness. Thanks to the iterative nature of the framework, our solution also provides a certain level of explanability, i.e. we can trace exactly how an image is preprocessed via a chain of transformations. In summary, we believe that this is a promising research approach to address the problem of automating data preprocessing. Future work should aim at addressing continuous actions, transformations that require memorization, and demonstrating the framework on other types of data such as text or time series.

\end{document}